\documentclass[a4paper,fleqn]{cas-dc}

\usepackage[authoryear]{natbib}

\def\tsc#1{\csdef{#1}{\textsc{\lowercase{#1}}\xspace}}
\tsc{WGM}
\tsc{QE}
\tsc{EP}
\tsc{PMS}
\tsc{BEC}
\tsc{DE}
\begin{document}
\let\WriteBookmarks\relax
\def\floatpagepagefraction{1}
\def\textpagefraction{.001}

\shorttitle{Mastering Nordschleife}

\shortauthors{Boettinger \& Klotz}

\title [mode = title]{Mastering Nordschleife - A comprehensive race simulation for AI strategy decision-making in motorsports}           
%\tnotemark[1,2]

%\tnotetext[1]{This document is the results of the research project funded by the National Science Foundation.}

%\tnotetext[2]{The second title footnote which is a longer text matter to fill through the whole text width and overflow into another line in the footnotes area of the first page.}

%    style=chinese,
%    auid=000,
%    bioid=1,
%    prefix=Sir,
%     orcid=0000-0000-0000-0000,
\author[1]{Max Boettinger}[type=editor,
                        auid=000,
                        orcid=0009-0007-2590-1963,
                        bioid=1
                        ]

%\cormark[1]


\ead{boettinger@hdm-stuttgart.de}

%\ead[url]{hdm-stuttgart.de}

%\credit{Conceptualization of this study, Methodology, Software}

\affiliation[1]{organization={Stuttgart Media University, Institute for Applied Artificial Intelligence},
    addressline={Nobelstr. 10}, 
    city={70569 Stuttgart},
    % citysep={}, % Uncomment if no comma needed between city and postcode
    % postcode={70569}, 
    country={Germany}}

\author[1]{David Klotz}[orcid=0000-0001-8322-0911]
%\orcid=0000-0001-8322-0911
%\cormark[1]
%\fnmark[1,3]
\ead{klotzd@hdm-stuttgart.de}
%\orcid=0000-0001-8322-0911
%\ead[URL]{hdm-stuttgart.de}

%\cortext[cor2]{Principal corresponding author}

%\fntext[fn1]{This is the first author footnote. but is common to third author as well.}
%\fntext[fn2]{Another author footnote, this is a very long footnote and it should be a really long footnote. But this footnote is not yet sufficiently long enough to make two lines of footnote text.}

%\nonumnote{This note has no numbers. In this work we demonstrate $a_b$
%  the formation Y\_1 of a new type of polariton on the interface
%  between a cuprous oxide slab and a polystirene micro-sphere placed
%  on the slab.
%  }

\begin{abstract}
In the realm of circuit motorsports, race strategy plays a pivotal role in determining race outcomes. This strategy focuses on the timing of pit stops, which are necessary due to fuel consumption and tire performance degradation. The objective of race strategy is to balance the advantages of pit stops, such as tire replacement and refueling, with the time loss incurred in the pit lane. Current race simulations, used to estimate the best possible race strategy, vary in granularity, modeling of probabilistic events, and require manual input for in-laps. This paper addresses these limitations by developing a novel simulation model tailored to GT racing and leveraging artificial intelligence to automate strategic decisions. By integrating the simulation with OpenAI's Gym framework, a reinforcement learning environment is created and an agent is trained. The study evaluates various hyperparameter configurations, observation spaces, and reward functions, drawing upon historical timing data from the 2020 Nürburgring Langstrecken Serie for empirical parameter validation.
The results demonstrate the potential of reinforcement learning for improving race strategy decision-making, as the trained agent makes sensible decisions regarding pit stop timing and refueling amounts. Key parameters, such as learning rate, decay rate and the number of episodes, are identified as crucial factors, while the combination of fuel mass and current race position proves most effective for policy development. The paper contributes to the broader application of reinforcement learning in race simulations and unlocks the potential for race strategy optimization beyond FIA Formula~1, specifically in the GT racing domain.
\end{abstract}

% \includegraphics{figs/grabs.pdf}

%\item Research highlights item 1
%\item Research highlights item 2
%\item Research highlights item 3

% Keywords

\maketitle

\section{Introduction}

The goal in circuit motorsports is to complete the race in the shortest possible time, maximizing the awarded points for the finishing position. Throughout the race, many factors can influence the result and informed strategic decision-making can give teams an edge over competitors. In this context, the timing and service choice of pit stops provides strategic potential. These stops become necessary, as fuel is being burned and tire performance degrades over time.

\subsection{Race strategy}
Determining the right amount of propellant and fitting the right tire compound at the right time can greatly impact a driver’s pace. Pit stops also have a drawback in the form of a time loss, as speed limits in the pit lane and the time needed to service the car allow opponents a chance to pass while on track. \cite{heilmeierRaceSimulationStrategy2018} therefore define race strategy as "weighting the costs and benefits of a pit stop”.
The timing can be influenced by both the driver---through a cautious driving style and thus less fuel consumption and tire stress--- and external influences like on-track accidents which might lead to speed limits induced by race control.
All these concepts provide the opportunity for tactical considerations. For example, a widely used tactical option is the so-called \emph{undercut}. It can be performed if the time delta between two cars is under a track-specific threshold, often around 1~second. If the trailing car decides to pit for fresh tires, it can use the performance difference of new tires after leaving the pits to pass, as the leading car must complete at least one lap on older and thus less performing tires.
Also, pitting in an active Safety Car phase provides great benefits, as all cars on track must adhere to a speed limit which decreases the speed delta between cars in the pit lane and cars not performing a pit stop, thus reducing the relative time cost of a pit stop.

\subsection{Race simulations}

\begin{figure}
	\centering
		\includegraphics[width=\linewidth]{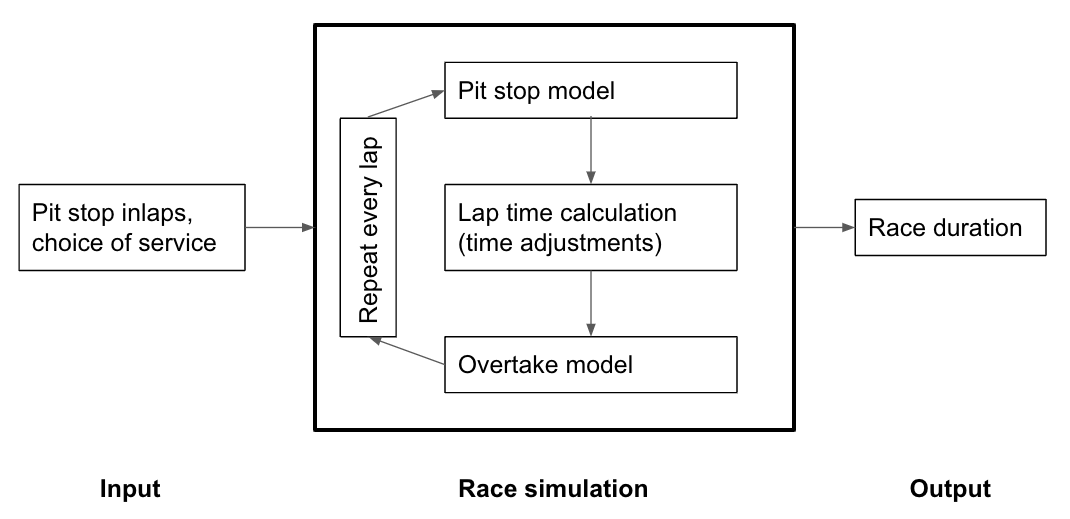}
	\caption{Race simulation structure.}
	\label{fig:race_simulation_structure}
\end{figure}

To prepare for the aforementioned complexity of race strategy decision-making, teams rely on in-depth simulations before and during races. As described by \cite{bekkerPlanningFormulaOne2009} and \cite{heilmeierVirtualStrategyEngineer2020} consequently, these simulations are used to evaluate the impact of strategic decisions on the resulting net race time. A common structure, modeled after \cite{heilmeierRaceSimulationStrategy2018}, is presented in figure~\ref{fig:race_simulation_structure} and shows that these simulations take input in the form of pit stop timing and desired actions for any given stop. This represents an improvement on the work by \cite{bekkerPlanningFormulaOne2009}, as many more input parameters were needed for their simulation.

Most simulations proposed in prior literature, such as \cite{bekkerPlanningFormulaOne2009} \cite{heilmeierRaceSimulationStrategy2018} \cite{heilmeierVirtualStrategyEngineer2020}, use a lap-wise discretization. In this approach, the race is simulated by calculating the lap times of all cars and applying sub-models to account for influencing factors. These range from modeling pit stops, overtakes, and race starts to yellow flags and safety car phases. The accumulated lap times result in a race time, which represents the race outcome.

\subsection{Automation of race simulations}
The most pressing shortcoming of basic race simulations, which limits both the amount of possible simulation runs as well as the predictive quality of those, is the requirement for manual input of pit stop decisions and timing for all simulated cars \cite{heilmeierVirtualStrategyEngineer2020}.

One solution for this problem, first proposed by \cite{heilmeierApplicationMonteCarlo2020}, is the use of deep neural networks. In their work, the authors trained two separate networks which decide whether to pit and which tire compound to fit, respectively. The results were promising, and the algorithm made sensible decisions based on the current race situation. \cite{liuFormulaERaceStrategy2020} described a different approach to replace the traditional lap time simulations using artificial neural networks and Monte Carlo tree search. The work focused on FIA Formula-E racing, where time-critical decisions have to be made for energy management. The resulting network demonstrated a decent capability for decision-making solutions for pre-race planning and reaction to in-race events. Their approach also proved to be much faster than other traditional simulation approaches.
The authors further expanded on their findings \cite{liuFormulaERaceStrategy2021} by applying Deep Deterministic Policy Gradient reinforcement learning. Here, the problem was formulated as a Markov decision process with a hybrid-type action space. The reward shaping was also implemented in a way to support both continuous and discrete components with soft boundary violation penalties, which was found to significantly improve the performance. Further, this proposal showed superior performance compared to the previous Monte Carlo Tree Search method.

This demonstrates the potential of using artificial intelligence, and specifically reinforcement learning, for improvements in race strategy decision-making. Additionally, the current state of research is limited to only the FIA Formula1 regulation, with their models not being applicable to other series, such as GT (Gran Turismo). Here, races are usually longer, with many focusing on the endurance of both man and machine. In contrast to the purpose-built F1 cars, GT races feature cars built upon their respective road versions. All of these differences are neglected as of now. However, as a major portion of all global racing series is subject to GT motorsport regulations, expanding the application of reinforcement learning to race simulations beyond FIA Formula1 offers a wide potential for race strategy optimization.

%This work aims to build on this potential by proposing novel models for a GT racing simulation and applying reinforcement learning to improve race strategy decision-making for races following GT motorsport regulation.

\subsection{Research goal}

The objective of this paper is to capitalize on the potential of reinforcement learning in race strategy optimization and develop a novel simulation model tailored to GT racing, specifically on the Nürburgring Nordschleife. To achieve this, existing models are thoroughly analyzed, and appropriate solutions are adapted to suit the unique characteristics of GT racing. Moreover, a reinforcement learning agent is trained within this environment to make strategic decisions that enhance race performance. The effectiveness of the agent's decision-making is then compared to a baseline model to evaluate the proposed approach's efficacy and demonstrate the potential improvements in GT racing strategy. In addition, historical timing data from the 2020 Nürburgring Langstrecken Serie is utilized to derive empirical parameters for the models, ensuring a more accurate and realistic representation of the racing environment.

% 2. ?? Related work
\section{Related work}

\subsection{Basic race simulations}
Race simulations model an entire race, taking into account various factors as illustrated in figure~\ref{fig:race_simulation_structure}. The primary goal is to observe the impact of strategic decisions on the resulting race time and the expected finishing position, rather than the performance of individual components on a lap-wise level. The structure and functionality, as well as the granularity of race simulations, differ significantly in prior literature, but some common elements can be identified. These are presented in the following subsections and grouped by their respective area of modeling.

\subsubsection{Base lap time and track discretization}
The first work on race simulations for strategy evaluation was published by \cite{bekkerPlanningFormulaOne2009}. The authors propose a time-adjustment approach, where a base lap time is used to model the differing potential of each competing participant. It is assumed that this time represents the best achievable lap time by a car with minimum fuel load and new tires, perfect driver performance, and no impairment due to traffic or weather conditions. This base time is then increased by time penalties, which account for the different sub-models of the race simulation, resulting in a lap time for each participant. It is assumed that car dynamic influences are already accounted for in the base lap time for each car.

The race is thereby simulated in a lap-wise fashion, where the accumulated lap times represent the final race time. This race time represents the race result when compared to all other cars. \cite{heilmeierRaceSimulationStrategy2018} built upon this proposal and improved the approach with a more granular model of different aspects of a motorsport race, such as empirically derived factors for both tire degradation and fuel loss. \cite{sulstersSimulatingFormulaOne2018} also utilize a base lap time with added time penalties to determine a driver's lap time.

The simulation proposed by \cite{bekkerPlanningFormulaOne2009} further differs from both \cite{heilmeierRaceSimulationStrategy2018} and \cite{sulstersSimulatingFormulaOne2018} in the way the racetrack is modeled. \cite{bekkerPlanningFormulaOne2009} decided on a sector-wise discretization to parameterize the individual features of different racetrack segments accordingly. For this, they divided the track into approximately 40 sectors, each 150m in length. The race is simulated by cars moving from one sector to the next. The time a car needs to complete any given sector is calculated by multiplying the sector proportion of the complete lap distance by the base lap time. The accumulated sector times represent the lap time, to which the described time penalties of sub-models are now added.

Such an approach requires additional information as to how each sector affects tire degradation, fuel consumption, and the impact on sector times and overtaking possibility through described factors. The data for said parameterization is acquired through telemetry data. The availability of such a dataset is one reason \cite{heilmeierRaceSimulationStrategy2018} decided against this approach and opted for modeling the racetrack as a single entity. Furthermore, such sector-wise discretization has implications for the overtaking model, with \cite{bekkerPlanningFormulaOne2009} limiting overtaking to the end of a sector and only one car at a time. \cite{heilmeierRaceSimulationStrategy2018} criticize this decision, arguing that in a given situation, multiple cars could overtake a slower car. 

\subsubsection{Time-adjustments}
To model the influences of wear, driver and crew performance, as well as probabilistic factors, time penalties are added to the base lap time. This approach is utilized in most current publications. As these influences vary in simulation and model granularity, they are being grouped and an introduction is given on a factor-wise level to improve readability.

\textbf{Fuel mass.} The impact of fuel mass on lap times over the length of a stint is present in all reviewed papers. \cite{bekkerPlanningFormulaOne2009} perform the calculation on a sector-wise level. In this approach, the time penalty for a car's fuel mass is further discretized into steps of 10 kilograms. This penalty is then multiplied by the amount of fuel on board, and the impact of a sector on the complete lap time is calculated by multiplying the resulting penalty by the proportion of a sector. \cite{heilmeierRaceSimulationStrategy2018} calculate the time penalty by multiplying the current fuel amount of a car by the lap sensitivity for this factor. This sensitivity, as well as the amount of fuel consumption per lap, is typically derived from a lap time simulation with an accurate engine consumption map. This approach requires achieving an accurate factor parametrization for each track before simulating a race. \cite{sulstersSimulatingFormulaOne2018} build a linear regression model to determine the impact of fuel load on lap times. This model is run separately for each driver before the race, as both the driver performance and the base consumption of a car can vary. For this, the difference between the fastest qualifying lap and a race lap is taken into account, and the intercept represents the average difference.

\textbf{Tire degradation.}
\cite{tremlettOptimalTyreUsage2016} investigated the factor of tire usage for FIA Formula~1 races. As racing tires produce much larger forces than road tires, due to a stiffer carcass and softer elastomeric rubber compounds, accurately modeling this is of great importance. \cite{farroniPhysicalModellingTire2017} used a thermo racing tire (TRT) model to achieve this and found it to return fitting results. As such models are mostly applied in car dynamics research and require both large amounts of telemetry data as well as a lot of computational resources, they cannot be used for this work, albeit having great potential for further research. \cite{bekkerPlanningFormulaOne2009} disregarded this factor and the publication is thus exempt in this segment. This omission is also criticized by other authors, especially \cite{heilmeierVirtualStrategyEngineer2020}, as this is an integral part of car dynamics and thus race strategy. \cite{heilmeierRaceSimulationStrategy2018} propose both a linear and a logarithmic function to determine the time loss through tire degradation. This is due to the complex nature of this factor since a tire initially underperforms when fitted, as its core temperature is below the optimal window. When the optimal temperature is reached, tire performance peaks and the best possible times can be achieved. From there, tire performance decreases over its age and lap times increase again, which is known as tire degradation. \cite{heilmeierVirtualStrategyEngineer2020} achieved better results when applying the logarithmic function but highlighted the need for accurate data. If data is sparse, the simpler linear model can be applied. \cite{sulstersSimulatingFormulaOne2018} proposes a quadratic regression model to model the described performance curve of a tire. She also considers a simpler linear model if tire degradation can be assumed to be low on a given track. In addition, the regression model is run separately for each driver, as a driver's driving style, along with the car's design, impacts the amount of tire degradation. The model expects the age of a tire as input, as well as the residual lap time after applying the fuel model. The model also includes coefficients to represent the different tire compound characteristics of the F1 regulation.

\subsection{Probabilistic models}

Motorsport races are subject to many probabilistic influences which need to be modeled accurately, including (but not limited to) accidents and failures, weather, driver performance, and opponent decisions. This is achieved by advanced models, which are applied and represented through a resulting time penalty in the race simulation. Prior research has been conducted to integrate probabilistic aspects into the simulation, whilst modeling them as accurately as possible albeit their nature. The work by \cite{heilmeierApplicationMonteCarlo2020} provides a comprehensive overview of the current state of academic publications in this field. 

\textbf{Race start.}
Accurately modeling the race start is of great importance, as cars are close to each other and many position changes are common in this phase of the race. \cite{phillipsUncoveringFormulaOne2014} and \cite{salminentRaceSimulatorDownloadable2019} both use a probability distribution of position changes per drive, as well as translating the starting performance to a time delta. This is done to cope with edge cases, where multiple drivers are assigned the same position after the race start. \cite{heilmeierApplicationMonteCarlo2020} criticize this approach since the driver performance distribution is skewed, with drivers regularly starting on the front of the grid having fewer cars to overtake and thus would be assigned a worse starting position. The authors solve this by initializing a baseline starter (mean starting time to reach the finish line) and comparing each driver to this model. 

\textbf{Variability of lap time and pit stops.}
When analyzing real-world lap times, it can be observed that no driver always delivers a perfect lap, with the times scattered around a mean value. To account for this, the majority of models use a sample from a Gauss distribution with zero mean and driver-specific standard deviation \citep{phillipsUncoveringFormulaOne2014, salminentRaceSimulatorDownloadable2019, heilmeierApplicationMonteCarlo2020}. A linear regression model is used in a single model \citep{tulabandhulaTireChangesFresh2014} but this approach falls short in modeling the stochastic nature of the variability, as a fixed time adjustment is used. Pit-stop times also vary but are not normally distributed. As such, both \cite{phillipsUncoveringFormulaOne2014} and \cite{heilmeierApplicationMonteCarlo2020} use a log-logistic distribution. This is fitted for each team and a time adjustment for each pit stop is added accordingly.

\textbf{Accidents and failures.}
\cite{dhanvanthMachineLearningBasedAnalytical2022} propose a machine learning approach, to derive the likelihood of accidents based on features extracted through a correlation analysis. This approach takes into account factors such as weather and the specific circuit. However, due to the sparse data available for the NLS racing series, it was deemed unsuitable for this work. \cite{bekkerPlanningFormulaOne2009} utilize a driver-specific probability to determine, in each lap, whether a given driver retires from the race due to an accident or failure. \cite{phillipsUncoveringFormulaOne2014} and \cite{salminentRaceSimulatorDownloadable2019} build on this approach and add a lap dependency (since many accidents happen in the first lap, where cars are most close to each other). Since this approach does not accurately model drivers with no retirements in the data set, \cite{sulstersSimulatingFormulaOne2018} uses Bayesian inference of all recorded retirements to also account for those cars. \cite{heilmeierApplicationMonteCarlo2020} also use this solution and note, that no prior publication differentiates between accidents (drivers fault) and retirements (teams fault). Their model has adapted accordingly.

\textbf{Full Course Yellow and Safety Car.}
As all prior work focuses on the FIA Formula~1 regulation, full course yellow (FCY) is implemented through a Virtual Safety Car (VSC) or Safety Car (SC) by race control. Both \cite{sulstersSimulatingFormulaOne2018} and \cite{phillipsUncoveringFormulaOne2014} do not make this differentiation and model the Safety car for 5 and 6 laps respectively. Another approach is to model the run-up phase (cars catching up to the waiting Safety Car) with a 20 percent lap time increase and the following phase (cars following the Safety Car) with a 40 percent lap time increase \citep{phillipsUncoveringFormulaOne2014}. \cite{salminentRaceSimulatorDownloadable2019} is the only study that makes the distinction between VSC and SC, but criticizes all prior models for different reasons: Prior work assumes, that all VSC phases always start and end at the beginning of a lap, which is unrealistic. Also, a VSC phase begins at the same time for all drivers, regardless of where they are on track. This makes modeling difficult, as some cars might be already a lap down from the leading car. Further, the time delta for pit-stops under the (V)SC as well as the lap times should be much higher as modeled in the publications. For this, \cite{heilmeierApplicationMonteCarlo2020} propose a \textit{Ghost Safety Car} which is calculated for each driver individually and only affects the respective car.

Also, as accidents and failures are the (most common) reason for a (V)SC phase, \cite{heilmeierApplicationMonteCarlo2020} model them in conjunction. To overcome the limitation of a lap-wise discretization, whilst assuring that a VSC always applies to all drivers regardless of the current track position, these phases are calculated before the actual race simulation. 

\subsection{Automation of race simulations}
The most pressing shortcoming of the aforementioned simulations, which limit both the amount of possible simulation runs as well as the predictive quality of those, is the requirement for manual input of pit stop decisions and timing for all simulated cars. 

\subsubsection{Statistics and basic ML}
\cite{tulabandhulaTireChangesFresh2014} used statistics and basic Machine Learning techniques, such as linear regression, to build a decision-making tool for pit-stop timing at NASCAR races. The results seemed promising, albeit starkly abstracting real-world influences, especially probabilistic events. \cite{chooRealtimeDecisionMaking2015} conducted research into different features which can be used to determine position changes in motorsport races. Furthermore, track characteristics that influence tire wear and fuel consumption are evaluated, which are used to determine the outcome of tire changes and therefore the results of simulated races. For this, races of the 2012 Nascar season are used. The author found a strong correlation between the amount of tire wear on a given track and the decision to either change 2 or 4 tires respectively. The trained model comes to a similar conclusion when making strategy predictions. Furthermore, the performance of the pit crew impacts the outcome of the races. This factor is found to be of greater importance than the race progresses, with late pit stops having the highest impact on race results. The proposed machine learning model was able to successfully predict the position of a driver when deciding to pit for a specific type of tire compound. \cite{tulabandhulaTireChangesFresh2014} go on to describe the potential for future research by including more features in the prediction model. 

\subsubsection{Deep neural networks}
In the study conducted by \cite{pengRankPositionForecasting2021}, deep learning models are employed to analyze time-series sequence data characterized by high levels of uncertainty. The researchers emphasize the importance of decomposing the cause-effect relationship, as it proves to be a critical factor for the successful performance of rank forecasting. In comparison to other deep forecasting models, the authors demonstrate that a hybrid approach, which combines an encoder-decoder network with a separate Multilayer Perceptron (MLP) network, is capable of delivering probabilistic forecasting. This method effectively models pit stop events and rank positions in car racing scenarios. \cite{heilmeierVirtualStrategyEngineer2020} propose the use of Machine Learning (ML) methods to overcome the limitations of previous race simulations. For this, a Virtual Strategy Engineer (VSE) is established, which is called after each lap and decides on further actions. The authors hereby trained two neural networks, which are called sequentially, and decide whether a pit stop is a viable option in the current situation and, if a pit stop is taken, which tire compound should be fitted. The race simulation is parametrized based on FIA Formula~1 regulation but adoption of other regulations would be possible. The networks are trained by historic race data of Formula~1 seasons 2014-2019 with the authors explaining the relative stability of regulations in this time frame. These networks were found to be able to make reasonable decisions with fast processing times. Furthermore, strategic decisions are taken when FCY phases occur or undercut attempts are being made by competitors. On the other hand, the authors describe the lack of proactive strategic behavior by the VSE as well as the prediction quality, which could be improved by merging the two networks together and forming stronger connections between the neurons.

\cite{liuFormulaERaceStrategy2020} focus on energy management in FIA Formula~E racing, which is an essential aspect of racing under this regulation. Balancing position fights and energy management throughout the race is crucial for developing an effective race strategy. Consequently, the decision-making process primarily involves energy management, with battery charge and temperature being of significant interest. The model proposed by \cite{liuFormulaERaceStrategy2020} serves as foundational work for academic research in this field, replacing traditional lap-time simulations with ANN prediction models and Monte Carlo tree search techniques.
In this approach, decisions are made lap-wise, considering seven input features that describe the limits for energy consumption, the amount of coasting (using available kinetic energy without applying power), and other factors. One crucial factor is the potential for regenerating energy through braking, which is only possible under a specific battery temperature. The ANNs demonstrate the feasibility of replacing traditional lap-time simulations, as they reasonably consider factors such as coasting and reactions to environmental events (e.g., SC phases). Moreover, the ANNs produced more simulation results with fewer errors in less time compared to traditional simulations. \cite{liuFormulaERaceStrategy2020} further discuss potential improvements by considering more input features and optimizing the networks for faster computing times.

\subsubsection{Reinforcement learning}
\cite{liuFormulaERaceStrategy2021} further build on their prior research into energy management for FIA Formula-E racing by applying a distributed policy gradient reinforcement learning method. The publication features a hybrid-type action space, which allows for both discrete as well as continuous actions. For this, the problem is first formulated as a Markov Decision Problem (MDP) which allows only taking the current state into account when considering further actions. The training implements prioritized experience replay, which helps to stabilize the training and build on past decisions with a favorable outcome. The reward shaping allows for soft boundary penalties, which significantly improve the results. Hereby, the reward function is split into a step-wise and a terminal function. For each step, either a positive reward of 5 is given or, if the decision is found to be unfavorable, a negative reward of -5 is given. After each episode, the battery temperature and power level as well as the race time are used to calculate the terminal reward. The authors also discuss the different implementations of reinforcement learning, starting with Q-Tables which store the possible reward of a state-action pair in a matrix-like table. As the policy in this approach is to simply pick the action that yields the highest reward, it is not feasible for more complex problems. Furthermore, algorithms like Actor-Critic (A2C) improve on Q-Tables, but training times are high and costly for an environment with control problems, especially if these include stochastic elements. As this is the case with race simulations (eg. safety car phases), \cite{liuFormulaERaceStrategy2021} decided on using a policy gradient approach. It is found, that the soft shaping of rewards results in better race times, as the critic can more easily estimate the rewards and help the actor to better find solutions. Furthermore, the continuous action space was found to be superior to a deterministic approach.

% 2. Race simulation
\section{Proposed race simulation}

\subsection{Methodology}
The objective of this research paper is to propose a novel machine learning (ML)-based approach for optimizing race strategy decision-making in GT racing. As there remain unanswered questions in prior research, the potential for real-world adoption is significant, particularly in GT racing where the barrier to entry is lower. Thus, this work aims to contribute to the expansion of the field by exploring the feasibility and potential benefits of utilizing ML in GT racing strategy decision-making. For this, some unique aspects of the underlying GT regulation must be considered. The proposed simulation adopts the \textit{Nürburgring Langstreckenserie (NLS)}, a racing series with 10~annual events all taking place at the \textit{Nordschleife}. The uniqueness hereby stems from the track and the heterogeneous starting grid. The latter allows vastly different cars all racing on the track at the same time, each competing in their respective group. The former features a 24~kilometers long lap through mountainous forest terrain, coining the nickname \textit{green hell}. Further, the race start and pit stop differ from the racing series used in previous publications, and novel models accounting for this need to be proposed. 

\subsection{Simulation approach}
To fulfill the intended objectives, the simulation replicates a typical 4-hour NLS race consisting of 25 laps under the SP9 (GT3) class regulation. In this simulation, an RL agent competes against 15 opponents that are simulated with a simplified race strategy of four pit stops, driven by fuel consumption limitations. This strategy incorporates a \textit{splash and dash} tactic, wherein the mandatory standing time is waived in the final laps of a race, allowing for shorter pit stops and thus reduction of the overall race time. The goal for the agent is to optimize pit stop decisions, i.e. the agent decides in each lap based on the available information about its conditions (e.g. fuel) and the race (e.g. the agent's position) whether to make a pit stop or not. The agent's decision-making process is modeled with a lap-wise discretization, making strategic choices before commencing each lap. Its primary objective is to determine the optimal race strategy under normal circumstances, incorporating the utilization of the \textit{splash and dash} tactic. Additionally, the agent must react appropriately to race incidents such as accidents and the respective slow zones following them,
%\textit{Code60} Phases and implement the \textit{undercut} procedure
to achieve a race time that is close to optimal and, consequently, yields a favorable race outcome.

\subsection{Lap time calculation}
The basis of our simulation of a driver's overall NLS race time \(T\) is its composition as the sum of the different lap times \(T_i\) of the race, as displayed in equation~\ref{eq:total_racetime}, where \(i\) is the lap number and \(m\) is the total number of laps, with \(i,m \in \mathbb{N}, \text{ and } 1 \leq i \leq m\).

\begin{equation}
    \begin{split}
    T &= \sum_{i=1}^m T_i \\
    \end{split}
    \label{eq:total_racetime}
\end{equation}

As the goal of this work is to implement a race simulation tailored to the regulation of the NLS, some specifics need to be expanded on. First, due to the unique topology of the track, the decision was made to further discretize the lap into sectors as in the study of \cite{bekkerPlanningFormulaOne2009}. This creates the possibility to accurately parameterize individual aspects of each sector and thus makes the simulation more realistic. Equation~\ref{eq:lap_discret} shows the breakdown of the lap time \(T_i\) into the sum of the times for each sector \(T_{i,s}\), where \(s\) is the sector number and \(n\) is the total number of sectors, with \(s,n \in \mathbb{N}, \text{ and } 1 \leq s \leq n\).

\begin{equation}
    \begin{split}
    T_i &= \sum_{s=1}^n T_{i,s}
%    T_i &= \sum_{s=1}^n B_{i,s} + D_{i,s} + F_{i,s} + T_{i,s} + P_{i,s} + S_{i,s}
    \end{split}
\label{eq:lap_discret}
\end{equation}

The time \(T_{i,s}\) for passing sector \(s\) during lap \(i\) is calculated using a time-adjustment approach, as described in \cite{salminentRaceSimulatorDownloadable2019,heilmeierRaceSimulationStrategy2018}. In this approach, each sector \(s\) has a base time \(\hat{T}_s\) that represents the best possible time achievable by a driver for the sector under ideal conditions, i.e. with no traffic, fresh tires, and little fuel. The simulation then subtracts time penalties for tire degradation, fuel mass, and traffic, depending on the current state of the tires, fuel level, and the race position of the simulated car. In addition, specific penalties for the race start are applied only at the beginning of the race. Given that penalties depend on the current conditions of the simulated car, they have been defined as function \(t: C \rightarrow \mathbb{R}^+; c \mapsto t(c)\), \(c \in C\) where \(C\) is the set of all states the virtual race car can be in, considering its tire degradation, fuel mass, race position, and lap number, and \(c\) is the current condition for a driver when the penalty is calculated (see equation~\ref{eq:sector_base_and_penalties}).

\begin{equation}
%    T_i &= \sum_{s=1}^n B_{i,s} + D_{i,s} + F_{i,s} + T_{i,s} + P_{i,s} + S_{i,s}
%    t(c) = t_r(c) + t_f(c) + t_b(c) + t_s(c)
    T_{i,s} = \hat{T}_s - t(c)
    \label{eq:sector_base_and_penalties}
\end{equation}

The total penalty function \(t(c)\) is calculated as the sum of the specific penalty functions for tire degradation (\(t_r\)), fuel mass (\(t_f\)), time loss due to other cars (\(t_b\)), and the additional time loss during the start (\(t_s\)), as displayed in equation~\ref{eq:total_penalty}.
%\( \mathbb{R}_+ \times \mathbb{R}_+\), where If \(R\) is the set \(t(c) = t_r(c) + t_f(c) + t_o(c) + t_s(c)\).

\begin{equation}
    \begin{split}
    t(c) &= t_r(c) + t_f(c) + t_b(c) + t_s(c)
    \end{split}
\label{eq:total_penalty}
\end{equation}

%Here, a base sector time---which represents the best possible time achievable by a driver with no traffic, fresh tires, and little fuel---is increased by time penalties derived from empiric models and can be seen in equation~\ref{eq:base_racetime}. 

%Further, some regulatory aspects differ in such ways that novel models need to be established. One example is the race start which is not performed standing in boxes on the grid as in F1 races but by driving in pairs of 2 until the race director signals the green light. Another example is the heterogeneous starting grid which consists of vastly differing cars, all racing within their respective class whilst being on the same track simultaneously. This introduces the probabilistic factor of traffic impairment, which needs to be accounted for in the simulation. As with \cite{salminentRaceSimulatorDownloadable2019} and \cite{heilmeierRaceSimulationStrategy2018}, the proposed simulation uses a time-adjustment approach. 

%\begin{equation}
%\begin{split}
%T_{race} &= \sum T_{lap} \\ &= \sum T_{sector base} + P_{tire deg} + P_{fuel mass} + P_{traffic} + P_{pit stop} + P_{race start}
%\end{split}
%\label{eq:1}
%\end{equation}

The penalty functions are described in subsection~\ref{subsec:Time-adjustment-models}. The simulation continuously iterates over each sector in each lap and results in an aggregated race time for each car. Further stochastic models are applied for probabilistic events such as overtakes, accidents, and lap time variance due to traffic impairment.

\subsection{Time-adjustment models}
\label{subsec:Time-adjustment-models}
To empirically derive realistic model parameters, a dataset from the NLS 2020 season is analyzed, comprising lap-wise timing data for all participants. As mentioned earlier, the starting grid features cars with significant differences in speed and dynamics, each competing in their respective class. To avoid mixing characteristics across different car classes, e.g. comparing a Fiat to a Ferrari, the data analysis focuses on a single class to accurately parameterize the models. We selected the \emph{SP9} class (limited to \emph{GT3} cars) as it is the most common regulation for GT racing and is featured in numerous series, including Le Mans 24h. 

\subsubsection{Tire wear and fuel consumption}
During a stint, the rubber on tires wears away due to contact with the track's asphalt, causing a reduction in surface material that can absorb heat, leading to decreased performance in a phenomenon called \emph{tire degradation}. Modeling tire performance is intricate, as newly fitted tires initially underperform until they reach the optimal temperature window, after which they gradually degrade. \cite{sulstersSimulatingFormulaOne2018} assessed various approaches and determined that a logarithmic function was the most suitable, which has been adopted for this study.

%\begin{equation}
%\begin{split}
%P_{tire deg} = f_{deg1} * log(f_{deg2} * t_{age} + c_{lap})
%t_r(c) = 
%\end{split}
%\end{equation}

As for fuel consumption, a linear function can express the continuous reduction in fuel and hence weight. Whilst neglecting some minor influences, a constant factor was found to be accurate by \cite{sulstersSimulatingFormulaOne2018} and is consequently adapted by \cite{heilmeierApplicationMonteCarlo2020}.

%\begin{equation}
%\begin{split}
%P_{fuel mass} = f_{sector} * m_{fuel amount} + p_{empty tank}
%\end{split}
%\end{equation}

As this simulation further discretizes laps into sectors, an individual factor \(s\) is introduced to both functions. It accounts for specifics such as elevation changes and track curvature. 

\subsubsection{Race start}
The race start under NLS regulation is executed as a rolling start. Here, all cars perform an introduction lap with reduced speed and, on reaching the final corner, line up pairwise whilst adhering to a speed limit. Once the race director signals green, cars accelerate and are allowed to overtake only after passing the starting line. Because this practice vastly differs from F1 regulation, a novel model needs to be established. This is made difficult by the fact, that car performance between different manufacturers is neutered through the concept of balance of performance (BOP), where weight is added to cars making them equal. Further, the allowance of up to four drivers per car makes an individual starting-factor attribution almost impossible. This led to the decision to derive the starting performance from a Gaussian distribution curve. It is established by calculating the average time a car needs from race start to the end of the first sector for each grid spot over all 2020 races. A pseudo-random value is now drawn for each participant when simulating the race start. As with \cite{phillipsUncoveringFormulaOne2014} and further expanded on by \cite{heilmeierApplicationMonteCarlo2020}, the probabilistic starting performance is expressed in seconds to contradict the possibility of collision, when two cars are assigned the same position. If the model still results in cars ending the first sector with the exact same time, it defaults to the original grid position order.

\subsubsection{Pit stops}
Pit stops in NLS races differ from other series through a mandatory standing time. The length is thereby dynamic and dependent on the race progress, reducing over time and allowing unregulated stops in the last laps. This introduces great strategic potential and is actively used for such matters. A pit stop can be discretized into three phases and is hence modeled this way. First, a car enters the pit lane and drives to its respective box, whilst adhering to the speed limit. This is expressed as travel-time-in and replaces the time for the last sector in the current lap. On reaching the crew, the current standing time is derived from the regulation for applicable stops. If no standing time is required, a static free service time is used. As delays in servicing almost never occur and empirical data is sparse, the decision was made to exclude probability from this factor. The time needed for leaving the pits and accelerating to race speed is expressed as travel time out and, aggregated with standing time or free service time respectively, used as time for the first sector of the following lap.

%\begin{equation}
%\begin{split}
%P_{pit stop} = t_{pit lane entry} + t_{standing time} + \\ t_{service time} + t_{pit lane exit}
%\end{split}
%\end{equation}

\subsubsection{Multi-class traffic and Code60}
\begin{figure}
	\centering
		\includegraphics[width=\linewidth]{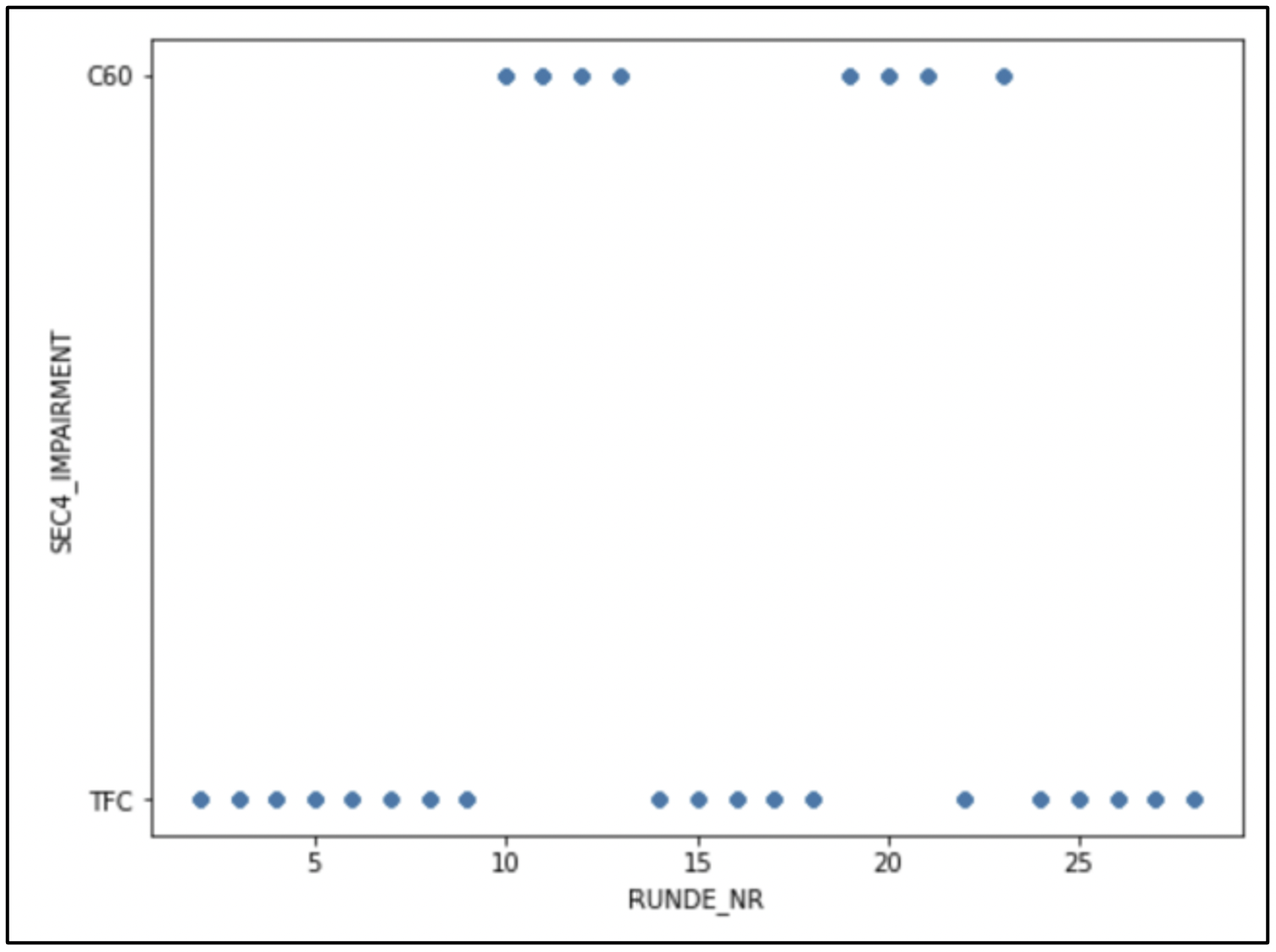}
	\caption{Sector 4 time variance classified as due to C60 or traffic.}
	\label{fig:sec_time_variance}
\end{figure}

NLS races feature a heterogeneous grid consisting of many different cars competing in individual classes whilst on the same track at the same time. As mentioned earlier, the proposed simulation models races of the SP9 GT class, whose cars belong to the fastest on track. This implies a great number of passes on vastly different and often slower cars, which again introduces the probabilistic factor of traffic impairment. The extent of this time variance can be observed in figure~\ref{fig:sec_time_variance}, mapping timing data on the y-axis and the lap count on the x-axis for sector 4 over a race. 

\begin{figure}
	\centering
		\includegraphics[width=\linewidth]{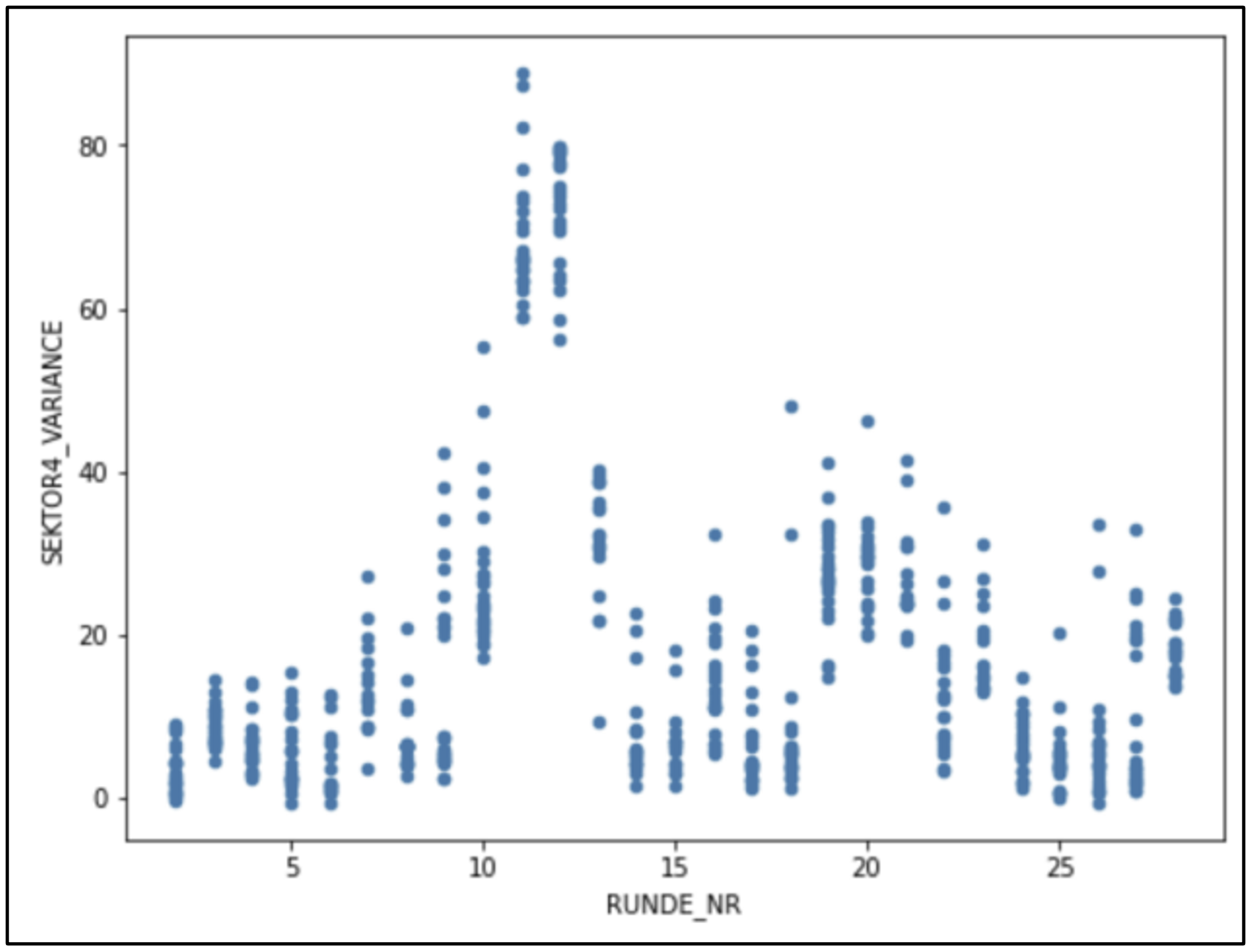}
	\caption{Time variance for given sector 4.}
	\label{fig:sector_time_variance}
\end{figure}

The lap-wise spread in sector times can be accounted to either being due to driver performance or traffic impairment. The spike on laps 11 and 12 is due to an active Code60 phase (C60), which is declared after a major incident took place and prohibits passing as well as limits the speed to 60 km/h in the respective sector. Such occurrences need to be excluded when modeling the probability of continuous traffic impairment. For this, a classification algorithm was used, and the results are shown in figure~\ref{fig:sector_time_variance}.

This approach was applied for each sector and the results were used to determine the probability of C60 phases in each sector. The topology of the respective sector was found to be influential, as sector~5, which mainly consists of a straight, and sector~4, consisting of windy and narrow track, differ greatly. Sector times classified as being due to traffic impairment were then used to build a probability distribution of time variance. The model now draws a value for each car when simulating a sector in every lap.

\subsubsection{Overtakes}
Previously proposed models cannot be used due to sector-wise discretization, with the exemption of \cite{bekkerPlanningFormulaOne2009}. The described approach was criticized by \cite{heilmeierRaceSimulationStrategy2018} and uses driver-specific probabilities which cannot be derived from our dataset. Hence, a novel solution is proposed. 

For this, the underlying dataset was used in order to establish empirically derived parameters of overtake probability per sector. These sector-specific parameters are necessary, as long straights enable drafting and thus passing with a greater time delta, while curvy sections require cars to follow each other closely. Also, the mean time delta between cars performing a successful overtake is used to decide whether an overtake should be simulated. If two cars are within the respective delta, a Gaussian probability distribution is used in order to model the overtake, with a time penalty for the overtaking car if the maneuver is unsuccessful.

% 3. Reinforcement Learning
\section{Application of reinforcement learning}
\label{sec:Appl_of_RL}
This study implemented Deep Q-Learning (DQL) using a Deep Q-Network (DQN) for optimizing strategy decisions. DQL builds on the concepts of Q-Learning and introduces deep neural networks to further improve the prediction accuracy as well as enable the application in more complex action and observation spaces. The need for such an integration of neural networks is due to the fact that in more complex state spaces, consisting of a greater amount of actions and observations, the iterative computation and updating of Q-Values becomes time and resource-critical. The concept of DQN hereby differs from traditional Q-Learning in that the value iteration for directly computing Q-Values and finding the optimal Q-Function is replaced by a function approximation, in order to estimate the optimal Q-Function. This approximation is achieved by training a deep neural network, which has been proven to be of great use for such tasks \cite{bohmerAutonomousLearningState2015}, hence enabling the field of Deep Q-Learning. The network receives the current state of an environment as input, requiring a direct mapping of input nodes to the size of the observation space. Furthermore, the network produces an output for every given action, which is the estimated Q-Value for taking this action in the given state. The objective of the network is, as with the traditional Q-Functions, to minimize the loss between the estimated Q-Value and the actually observed rewards. This is also used as an evaluation metric for the training later on.

\subsection{Action and observation space}
Integral parts of reinforcement learning are both action and observation spaces. These allow the agent to take any given action and receive an observation of the current environment state respectively. As both definitions greatly impact the resulting quality of decision-making by the agent, it has to be chosen accurately.

As explained earlier, the underlying regulation of the NLS racing series allows for refueling. Due to the track length, the limiting factor of stint length is, in contrast to other popular racing series, the fuel amount and not tire degradation. This has to be resembled in the action space, as the agent needs to learn when to pit and by how much the car should be fueled. This should allow the agent to both keep the car in the race (not running out of fuel) as well as take advantage of tactical possibilities such as yellow flags and the discussed over and undercuts. Further, the agent needs to understand and make use of the pit stop regulation, as later stops in the race no longer need to comply with mandatory standing times and thus are much quicker. These factors are displayed in table~\ref{tab:table-actionspace}. 

\begin{table}[width=.9\linewidth,cols=4,pos=h]
\caption{Reinforcement learning action space}\label{tab:table-actionspace}
\begin{tabular*}{\tblwidth}{@{} LLLL@{} }
\toprule
Action & Description\\
\midrule
0 & No pit stop, continue the race \\
1 & Pit and refuel for 4 laps \\
2 & Pit and refuel for 6 laps \\
3 & Pit and refuel for 8 laps \\
\bottomrule
\end{tabular*}
\end{table}

For allowing the agent to take the right decision in the current state, a meaningful observation space was needed. As the actions directly influence the amount of fuel in the car, this was chosen as the first observation. Because the overarching goal for the agent is to finish the race first, the current race position is also given as an observation in each step.

\subsection{Reward shaping}
As described earlier, the goal of an agent is to maximize its cumulative total reward, which is known as the reward hypothesis. The agent hereby does not have direct control over the environment and only receives feedback in form of a reward. What the agent can decide on, is what action to take in order to maximize its expected reward when transitioning into the next state. This is also known as a policy, which is defined as the probability of taking action \(a\) at time \(t\) when the agent is in the state \(s\) at time \(t\). The agent now tries to learn a mapping function from state to action, which can be used to estimate the possible reward for taking a given action. The policy an agent is following can also be described as behavior and directly dictates the actions that are being taken.

After taking an action and determining the next state of the environment, a step-wise reward is
being calculated and given to the agent. This work uses a reward space of \([-1,1]\) when calculating the step-wise reward, where a positive reward is assigned when the race position is ranked 4 or better in each step. A negative reward is given if the agent’s car reaches a tire degradation above a value of 90, as this results in both slower lap times and the possibility of retirement by reaching a critical tire degradation. Furthermore, a terminal reward is given based on the resulting end position. If the agent retires from the race due to high tire degradation, a terminal reward of -10 is assigned. As for the proposed simulation and articulated goal, ``race position'', ``race progression'', ``time delta to leading car'', and ``fuel level'' are most promising as reward signals for optimization towards the best strategy decisions, thus being evaluated in this work. 

\subsection{Hyper-parameter configuration}
The network used for the Q-Function approximation is a convolutional neural network and consists
of two input nodes, 50 dense layers with a relu activation layer, and a fully-connected output
layer. This output layer consists of four nodes, which directly represent the four possible actions that can be chosen by the agent. The weights of the network are initialized randomly, similar to the Q-Values in the Q-Table when using Q-Learning. In order to stabilize the training, and counteract training problems such as ``catastrophic forgetting'' \cite{kirkpatrickOvercomingCatastrophicForgetting2017}, experience replay is being applied. Hereby, the agent’s experience for each time step is saved in a replay memory. This can be expressed as 

\begin{equation}
\begin{split}
e_{t} = (s_{t}, a_{t}, r_{t+1}, s_{t+1})
\end{split}
\end{equation}

and consists of the respective state \(s_t\), the choice \(a_t\) taken by the agent, the assigned reward \(r_{t+1}\) for that action, and the resulting environment \(s_{t+1}\). This is done in order to dis-correlate consecutive sequential experiences that occur during training, which would lead to learning these correlations and hence inefficient training. Further hyper-parameter configurations have been evaluated, such as \textit{number of episodes}, \textit{replay buffer length}, \textit{batch size}, and the \textit{learning rate}. The results are discussed in the following chapter and the best configuration is chosen for validating the resulting network.

% 4. Results
\section{Results}

\subsection{Base line model}
\begin{figure}
	\centering
		\includegraphics[width=\linewidth]{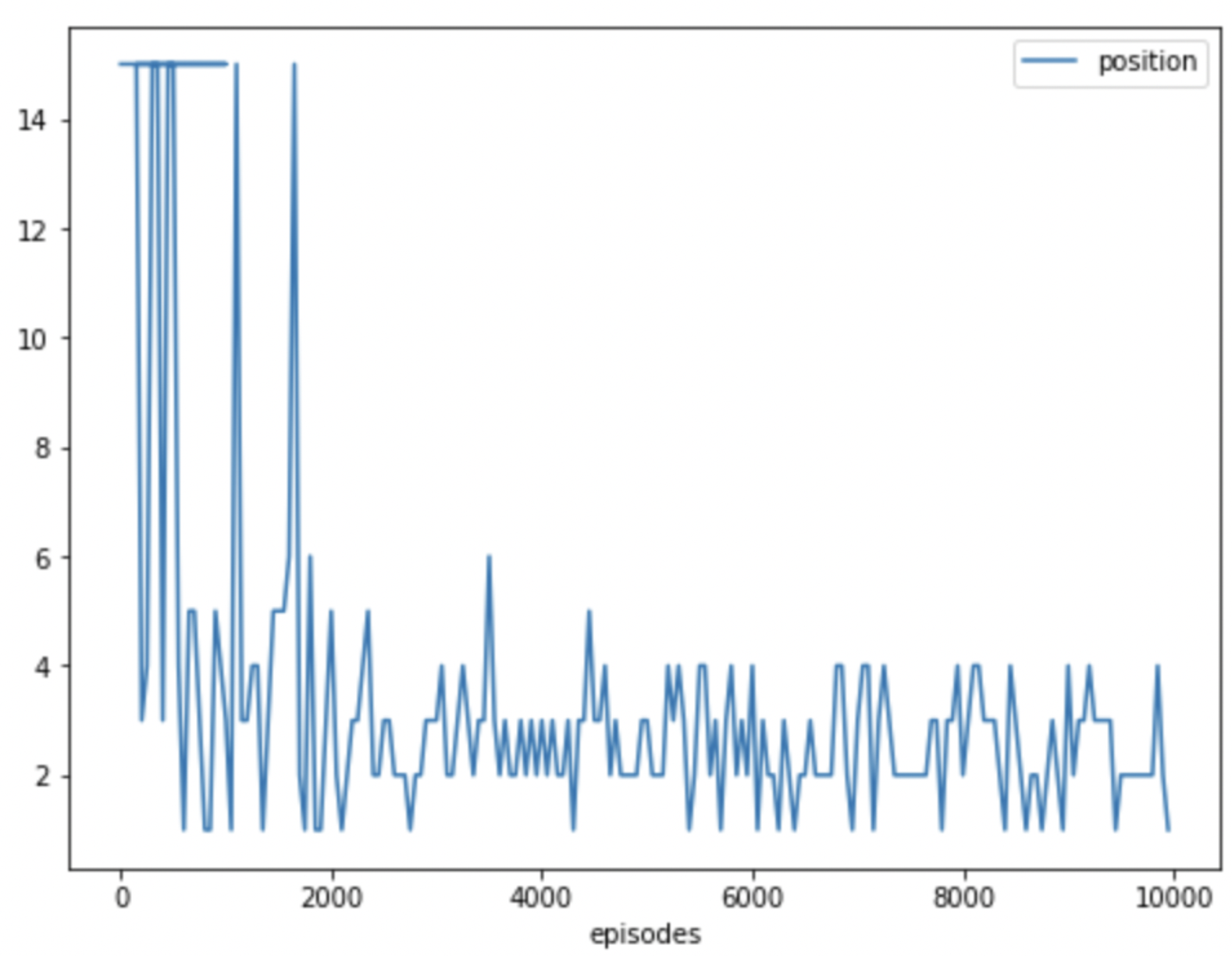}
	\caption{Initial Q-Learning race results.}
	\label{fig:result_qlearning}
\end{figure}

\begin{figure}
	\centering
		\includegraphics[width=\linewidth]{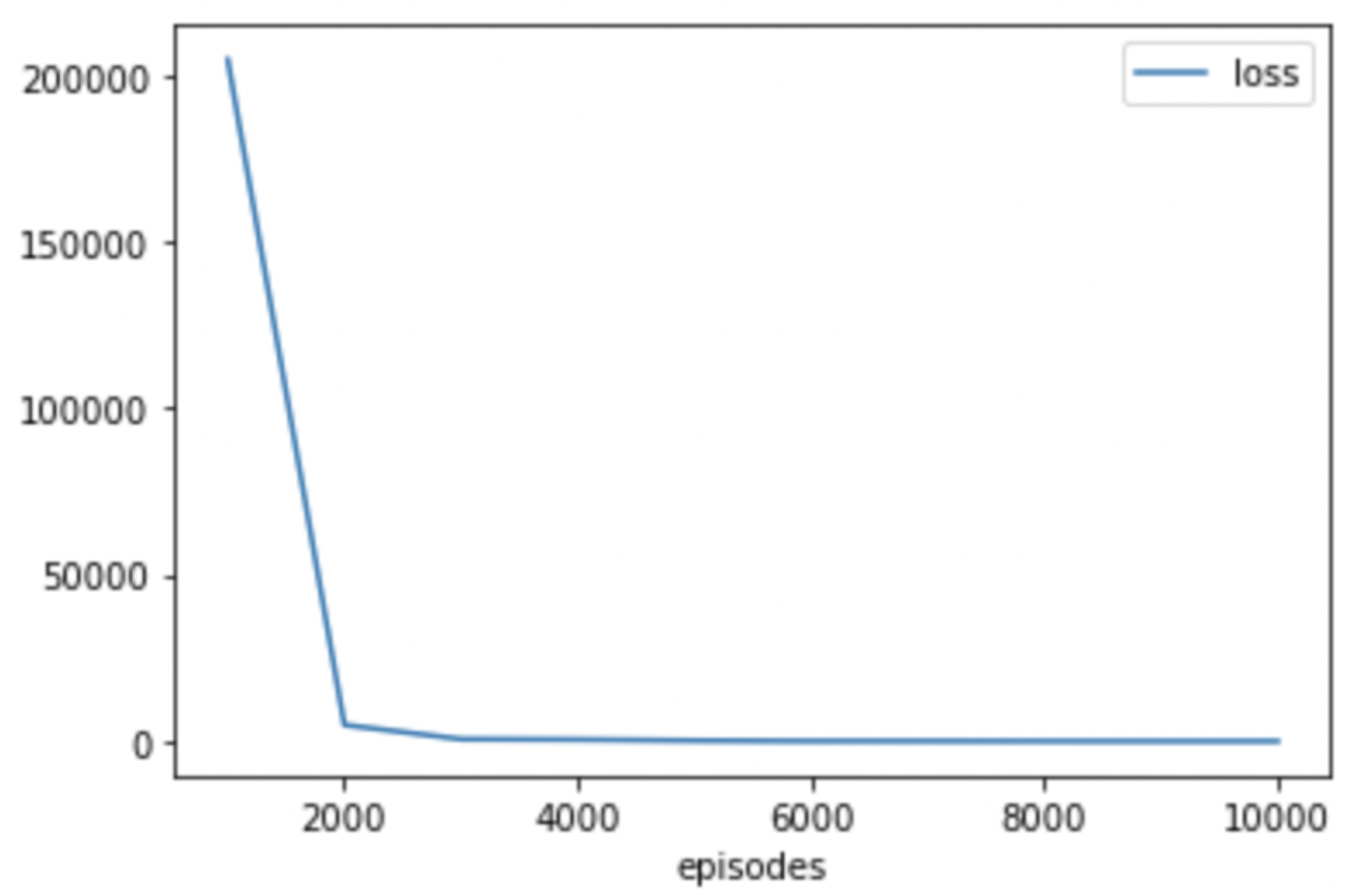}
	\caption{Initial Q-Learning loss.}
	\label{fig:results_dqn_loss}
\end{figure}

As the state space of the established environment is of considerable size, the use of Q-Learning has limited potential. Therefore, this algorithm was only used as a baseline model for comparison to the performance of the DQN. The observation space consisted of the current race position as well as the current tire degradation factor. The action space was configured as outlined in section~\ref{sec:Appl_of_RL}, which was also the case for the reward function. When running the training as described, the resulting loss per every 50th episode can be seen in figure~\ref{fig:results_dqn_loss}. As the loss is stabilizing, the reward is increased over time. This correlates with the achieved end positions which get lower over the training duration and can be observed in \ref{fig:result_qlearning}. This behavior can be explained by the simplistic policy applied in Q-Learning which starts off by choosing random actions (exploration) and filling the Q-Learning. After that, the algorithm begins to select the actions with the highest associated reward (exploitation), thus converging and achieving repeated race wins.

\subsection{DQN model}
As for the DQN model, different hyperparameter configurations were examined. For this, the same observation space and reward function as with the baseline Q-Learning model have
been used for the training. The initial results for loss and average finishing positions can be seen in figure~\ref{fig:results_dqn_loss} and figure~\ref{fig:result_qlearning}.

Results of the first variation show, that training quickly stabilized when using a learning rate of 0.001, but the achieved reward stagnated at the same time. Such a behavior can be explained by either the agent stagnating on a plateau, stemming from early exploitation and thus neglecting a possible better strategy, or by insufficient reward shaping. Furthermore, deciding on 10,000 episodes might not be enough in order to calculate the correct wights
for the network. Either of these factors could explain the results of the second variation of the DQN training. Hereby, the learning rate was increased to 0.01 and the episodes increased to 50,000. The results still lack sturdiness, as an early convergence in both average reward and loss indicates that a local maximum was found and was not overcome. The agent was able to successfully complete the race distance but continued to be classified last. This leads to a reward of around 0, still avoiding the negative rewards of being retired or passed several times, but not the desired outcome. 

Better results have been achieved with an increased amount of episodes. These have been doubled to 100,000 in the third variation of hyper-parameter evaluation. Hereby, greater variance and loss were observed in the first part of training, with race results varying greatly from wins to being classified last by a significant margin. But when reaching the 70,000th episode, a convergence was observed and the race results stabilized, with consistent first place
finishes. 

\begin{figure}
	\centering
		\includegraphics[width=\linewidth]{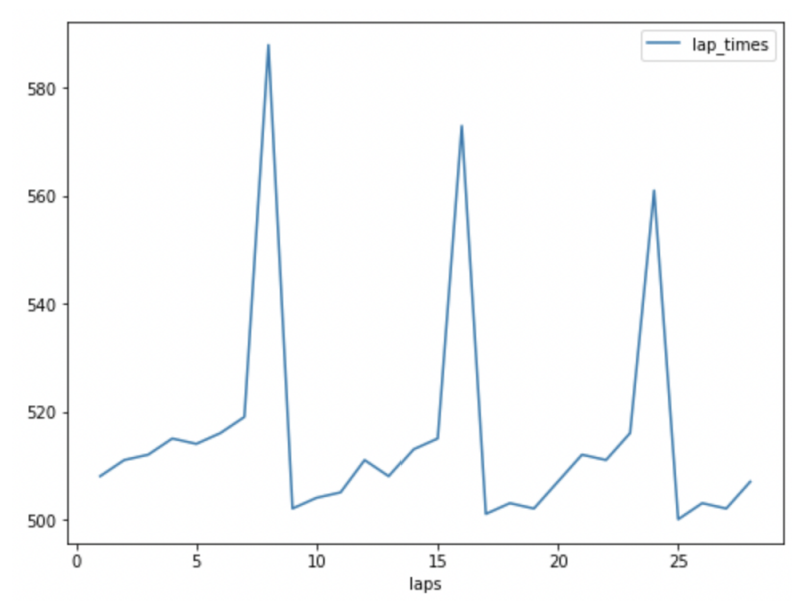}
	\caption{Optimal race strategy chosen by the network}
	\label{fig:dqlearning_best_strategy}
\end{figure}

Upon evaluating the outcomes of the agent using this policy, the most favorable race strategy is identified. This aligns with strategies applied in actual four-hour NLS races, where an initial stint of 8 laps is followed by two more stints of 8 laps each. For the final laps, the advantage of unregulated standing times is leveraged, and refueling is adjusted accordingly. This is illustrated in figure~\ref{fig:dqlearning_best_strategy}, where the agent is found to execute this precise procedure.

% 6. Conclusion
\section{Conclusion}
This paper extends prior research in the field of automated race strategy decision-making as well as modeling deterministic and probabilistic factors of race simulations. A comprehensive simulation of GT races---specific to the NLS regulation---was implemented. Empirical factors were derived from real-world data and used to accurately parameterize models. As the regulation of the underlying racing series vastly differs from previous publications, novel models were established. Models for tire degradation and fuel consumption proved accurate and were adopted successfully. As for the newly proposed approaches, some adjustments to the set goals had to be made. The quality of the underlying dataset proved worse than expected and deriving model parameters proved challenging. Nonetheless, the simulation was found to accurately represent a real-world GT race on the Nürburgring Nordschleife, which builds the groundwork for further academic work in this underrepresented field.

The simulation was used as an environment for training a reinforcement learning agent. Different hyper-parameter configurations have been evaluated and the most promising configuration validated against a simplistic Q-Learning baseline model. We found significant improvements over this baseline model with the agent consistently choosing appropriate actions in the respective race context.

Yet, research about the probabilistic factors of GT-starts, multi-class traffic, and topology-based parametrization is sparse. Thus, further research into these factors is required and could improve the sturdiness of the implemented simulation. In addition, the application of reinforcement learning for race strategy decision-making is still in its early phases and this work solidifies its prospects. Also, the concept of ``self-play'' could be of great potential for applications in such environments. Hereby, the training of an agent is enhanced by ``playing'' against earlier versions of itself, thus rendering the static participants of race simulations unnecessary. By putting the actions of the network in direct contest with each other, better and more robust policies are expected as outcomes.

All in all, the automation of race strategy decision-making as well as the accurate modeling of
sturdy race simulations with regard to probabilistic factors and the application of these concepts to real-world racing has great potential. These could change the workflow and real-time decision-making of race engineers and thus influence the outcome of future motorsport races.

\bibliographystyle{model1-num-names}
%\bibliographystyle{cas-model2-names}

\bibliography{./article.bib}

\begin{thebibliography}{17}
\expandafter\ifx\csname natexlab\endcsname\relax\def\natexlab#1{#1}\fi
\providecommand{\url}[1]{\texttt{#1}}
\providecommand{\href}[2]{#2}
\providecommand{\path}[1]{#1}
\providecommand{\DOIprefix}{doi:}
\providecommand{\ArXivprefix}{arXiv:}
\providecommand{\URLprefix}{URL: }
\providecommand{\Pubmedprefix}{pmid:}
\providecommand{\doi}[1]{\href{http://dx.doi.org/#1}{\path{#1}}}
\providecommand{\Pubmed}[1]{\href{pmid:#1}{\path{#1}}}
\providecommand{\bibinfo}[2]{#2}
\ifx\xfnm\relax \def\xfnm[#1]{\unskip,\space#1}\fi
%Type = Inproceedings
\bibitem[{Heilmeier et~al.(2018)Heilmeier, Graf, and
  Lienkamp}]{heilmeierRaceSimulationStrategy2018}
\bibinfo{author}{A.~Heilmeier}, \bibinfo{author}{M.~Graf},
  \bibinfo{author}{M.~Lienkamp},
\newblock \bibinfo{title}{A {{Race Simulation}} for {{Strategy Decisions}} in
  {{Circuit Motorsports}}},
\newblock in: \bibinfo{booktitle}{2018 21st {{International Conference}} on
  {{Intelligent Transportation Systems}} ({{ITSC}})}, \bibinfo{year}{2018}, pp.
  \bibinfo{pages}{2986--2993}. \DOIprefix\doi{10.1109/ITSC.2018.8570012}.
%Type = Article
\bibitem[{Bekker and Lotz(2009)}]{bekkerPlanningFormulaOne2009}
\bibinfo{author}{J.~Bekker}, \bibinfo{author}{W.~Lotz},
\newblock \bibinfo{title}{Planning {{Formula One}} race strategies using
  discrete-event simulation},
\newblock \bibinfo{journal}{Journal of the Operational Research Society}
  \bibinfo{volume}{60} (\bibinfo{year}{2009}) \bibinfo{pages}{952--961}.
%Type = Article
\bibitem[{Heilmeier et~al.(2020{\natexlab{a}})Heilmeier, Thomaser, Graf, and
  Betz}]{heilmeierVirtualStrategyEngineer2020}
\bibinfo{author}{A.~Heilmeier}, \bibinfo{author}{A.~Thomaser},
  \bibinfo{author}{M.~Graf}, \bibinfo{author}{J.~Betz},
\newblock \bibinfo{title}{Virtual {{Strategy Engineer}}: {{Using Artificial
  Neural Networks}} for {{Making Race Strategy Decisions}} in {{Circuit
  Motorsport}}},
\newblock \bibinfo{journal}{Applied Sciences} \bibinfo{volume}{10}
  (\bibinfo{year}{2020}{\natexlab{a}}) \bibinfo{pages}{7805}.
%Type = Article
\bibitem[{Heilmeier et~al.(2020{\natexlab{b}})Heilmeier, Graf, Betz, and
  Lienkamp}]{heilmeierApplicationMonteCarlo2020}
\bibinfo{author}{A.~Heilmeier}, \bibinfo{author}{M.~Graf},
  \bibinfo{author}{J.~Betz}, \bibinfo{author}{M.~Lienkamp},
\newblock \bibinfo{title}{Application of {{Monte Carlo Methods}} to {{Consider
  Probabilistic Effects}} in a {{Race Simulation}} for {{Circuit Motorsport}}},
\newblock \bibinfo{journal}{Applied Sciences} \bibinfo{volume}{10}
  (\bibinfo{year}{2020}{\natexlab{b}}) \bibinfo{pages}{4229}.
%Type = Article
\bibitem[{Liu and Fotouhi(2020)}]{liuFormulaERaceStrategy2020}
\bibinfo{author}{X.~Liu}, \bibinfo{author}{A.~Fotouhi},
\newblock \bibinfo{title}{Formula-{{E}} race strategy development using
  artificial neural networks and {{Monte Carlo}} tree search},
\newblock \bibinfo{journal}{Neural Computing and Applications}
  \bibinfo{volume}{32} (\bibinfo{year}{2020}) \bibinfo{pages}{15191--15207}.
%Type = Article
\bibitem[{Liu et~al.(2021)Liu, Fotouhi, and
  Auger}]{liuFormulaERaceStrategy2021}
\bibinfo{author}{X.~Liu}, \bibinfo{author}{A.~Fotouhi}, \bibinfo{author}{D.~J.
  Auger},
\newblock \bibinfo{title}{Formula-{{E}} race strategy development using
  distributed policy gradient reinforcement learning},
\newblock \bibinfo{journal}{Knowledge-Based Systems} \bibinfo{volume}{216}
  (\bibinfo{year}{2021}) \bibinfo{pages}{106781}.
%Type = Misc
\bibitem[{Sulsters(2018)}]{sulstersSimulatingFormulaOne2018}
\bibinfo{author}{C.~Sulsters}, \bibinfo{title}{Simulating {{Formula One Race
  Strategies}}},
  \bibinfo{howpublished}{https://www.semanticscholar.org/paper/Simulating-Formula-One-Race-Strategies-Sulsters/659f14b5beafa7a8e5f5eaebdb9b0fdfb85a0ebe},
  \bibinfo{year}{2018}.
%Type = Article
\bibitem[{Tremlett and Limebeer(2016)}]{tremlettOptimalTyreUsage2016}
\bibinfo{author}{A.~J. Tremlett}, \bibinfo{author}{D.~J.~N. Limebeer},
\newblock \bibinfo{title}{Optimal tyre usage for a {{Formula One}} car},
\newblock \bibinfo{journal}{Vehicle System Dynamics} \bibinfo{volume}{54}
  (\bibinfo{year}{2016}) \bibinfo{pages}{1448--1473}.
%Type = Article
\bibitem[{Farroni et~al.(2017)Farroni, Sakhnevych, and
  Timpone}]{farroniPhysicalModellingTire2017}
\bibinfo{author}{F.~Farroni}, \bibinfo{author}{A.~Sakhnevych},
  \bibinfo{author}{F.~Timpone},
\newblock \bibinfo{title}{Physical modelling of tire wear for the analysis of
  the influence of thermal and frictional effects on vehicle performance},
\newblock \bibinfo{journal}{Proceedings of the Institution of Mechanical
  Engineers, Part L: Journal of Materials: Design and Applications}
  \bibinfo{volume}{231} (\bibinfo{year}{2017}) \bibinfo{pages}{151--161}.
%Type = Article
\bibitem[{Phillips(2014)}]{phillipsUncoveringFormulaOne2014}
\bibinfo{author}{A.~J.~K. Phillips},
\newblock \bibinfo{title}{Uncovering {{Formula One}} driver performances from
  1950 to 2013 by adjusting for team and competition effects},
\newblock \bibinfo{journal}{Journal of Quantitative Analysis in Sports}
  \bibinfo{volume}{10} (\bibinfo{year}{2014}) \bibinfo{pages}{261--278}.
%Type = Misc
\bibitem[{Salminen(2019)}]{salminentRaceSimulatorDownloadable2019}
\bibinfo{author}{B.~E. Salminen, T}, \bibinfo{title}{Race {{Simulator}}:
  {{Downloadable R Program Code}}}, \bibinfo{year}{2019}.
%Type = Article
\bibitem[{Tulabandhula and Rudin(2014)}]{tulabandhulaTireChangesFresh2014}
\bibinfo{author}{T.~Tulabandhula}, \bibinfo{author}{C.~Rudin},
\newblock \bibinfo{title}{Tire {{Changes}}, {{Fresh Air}}, and {{Yellow
  Flags}}: {{Challenges}} in {{Predictive Analytics}} for {{Professional
  Racing}}},
\newblock \bibinfo{journal}{Big Data} \bibinfo{volume}{2}
  (\bibinfo{year}{2014}) \bibinfo{pages}{97--112}.
%Type = Inproceedings
\bibitem[{Dhanvanth et~al.(2022)Dhanvanth, Rajesh, Samyukth, and
  Jeyakumar}]{dhanvanthMachineLearningBasedAnalytical2022}
\bibinfo{author}{S.~Dhanvanth}, \bibinfo{author}{R.~Rajesh},
  \bibinfo{author}{S.~S. Samyukth}, \bibinfo{author}{G.~Jeyakumar},
\newblock \bibinfo{title}{Machine {{Learning-Based Analytical}} and
  {{Predictive Study}} on {{Formula}} 1 and {{Its Safety}}},
\newblock in: \bibinfo{editor}{A.~Tomar}, \bibinfo{editor}{H.~Malik},
  \bibinfo{editor}{P.~Kumar}, \bibinfo{editor}{A.~Iqbal} (Eds.),
  \bibinfo{booktitle}{Proceedings of 3rd {{International Conference}} on
  {{Machine Learning}}, {{Advances}} in {{Computing}}, {{Renewable Energy}} and
  {{Communication}}}, Lecture {{Notes}} in {{Electrical Engineering}},
  \bibinfo{publisher}{{Springer Nature}}, \bibinfo{address}{{Singapore}},
  \bibinfo{year}{2022}, pp. \bibinfo{pages}{257--266}.
  \DOIprefix\doi{10.1007/978-981-19-2828-4_25}.
%Type = Phdthesis
\bibitem[{Choo(2015)}]{chooRealtimeDecisionMaking2015}
\bibinfo{author}{C.~L.~W. Choo}, \bibinfo{title}{Real-Time Decision Making in
  Motorsports : Analytics for Improving Professional Car Race Strategy},
  \bibinfo{type}{Thesis}, Massachusetts Institute of Technology,
  \bibinfo{year}{2015}.
%Type = Inproceedings
\bibitem[{Peng et~al.(2021)Peng, Li, Akkas, Araki, Yoshiyuki, and
  Qiu}]{pengRankPositionForecasting2021}
\bibinfo{author}{B.~Peng}, \bibinfo{author}{J.~Li}, \bibinfo{author}{S.~Akkas},
  \bibinfo{author}{T.~Araki}, \bibinfo{author}{O.~Yoshiyuki},
  \bibinfo{author}{J.~Qiu},
\newblock \bibinfo{title}{Rank {{Position Forecasting}} in {{Car Racing}}},
\newblock in: \bibinfo{booktitle}{2021 {{IEEE International Parallel}} and
  {{Distributed Processing Symposium}} ({{IPDPS}})}, \bibinfo{year}{2021}, pp.
  \bibinfo{pages}{724--733}. \DOIprefix\doi{10.1109/IPDPS49936.2021.00082}.
%Type = Article
\bibitem[{B{\"o}hmer et~al.(2015)B{\"o}hmer, Springenberg, Boedecker,
  Riedmiller, and Obermayer}]{bohmerAutonomousLearningState2015}
\bibinfo{author}{W.~B{\"o}hmer}, \bibinfo{author}{J.~T. Springenberg},
  \bibinfo{author}{J.~Boedecker}, \bibinfo{author}{M.~Riedmiller},
  \bibinfo{author}{K.~Obermayer},
\newblock \bibinfo{title}{Autonomous {{Learning}} of {{State Representations}}
  for {{Control}}: {{An Emerging Field Aims}} to {{Autonomously Learn State
  Representations}} for {{Reinforcement Learning Agents}} from {{Their
  Real-World Sensor Observations}}},
\newblock \bibinfo{journal}{KI - K\"unstliche Intelligenz} \bibinfo{volume}{29}
  (\bibinfo{year}{2015}) \bibinfo{pages}{353--362}.
%Type = Article
\bibitem[{Kirkpatrick et~al.(2017)Kirkpatrick, Pascanu, Rabinowitz, Veness,
  Desjardins, Rusu, Milan, Quan, Ramalho, {Grabska-Barwinska}, Hassabis,
  Clopath, Kumaran, and
  Hadsell}]{kirkpatrickOvercomingCatastrophicForgetting2017}
\bibinfo{author}{J.~Kirkpatrick}, \bibinfo{author}{R.~Pascanu},
  \bibinfo{author}{N.~Rabinowitz}, \bibinfo{author}{J.~Veness},
  \bibinfo{author}{G.~Desjardins}, \bibinfo{author}{A.~A. Rusu},
  \bibinfo{author}{K.~Milan}, \bibinfo{author}{J.~Quan},
  \bibinfo{author}{T.~Ramalho}, \bibinfo{author}{A.~{Grabska-Barwinska}},
  \bibinfo{author}{D.~Hassabis}, \bibinfo{author}{C.~Clopath},
  \bibinfo{author}{D.~Kumaran}, \bibinfo{author}{R.~Hadsell},
\newblock \bibinfo{title}{Overcoming catastrophic forgetting in neural
  networks},
\newblock \bibinfo{journal}{Proceedings of the National Academy of Sciences}
  \bibinfo{volume}{114} (\bibinfo{year}{2017}) \bibinfo{pages}{3521--3526}.

\end{thebibliography}


%Author biography without author photo.

%\bio{figs/pic1}
%Author biography with author photo.

\end{document}